\documentclass[10pt,twocolumn,letterpaper]{article}

\usepackage{cvpr}
\usepackage{times}
\usepackage{epsfig}
\usepackage{graphicx}
\usepackage{amsmath}
\usepackage{amssymb}
\usepackage{multirow}
\usepackage{epstopdf}
\usepackage[pagebackref=true,breaklinks=true,letterpaper=true,colorlinks,bookmarks=false]{hyperref}

\cvprfinalcopy % *** Uncomment this line for the final submission

\def\cvprPaperID{3446} % *** Enter the CVPR Paper ID here

\newcommand{\myparagraph}[1]{\vspace{3pt}\noindent{\bf #1}}
\newcommand*\samethanks[1][\value{footnote}]{\footnotemark[#1]}

\ifcvprfinal\fi
\begin{document}

%%%%%%%%% TITLE
\title{Parameter-Free Spatial Attention Network for Person Re-Identification}

% \author{\IEEEauthorblockN{Haoran Wang\IEEEauthorrefmark{1},
% Author Two\IEEEauthorrefmark{2}, Author Three\IEEEauthorrefmark{3} and
% Author Four\IEEEauthorrefmark{4}}
% \IEEEauthorblockA{Department of Whatever,
% Whichever University\\
% Wherever\\
% Email: \IEEEauthorrefmark{1}author.one@add.on.net,
% \IEEEauthorrefmark{2}author.two@add.on.net,
% \IEEEauthorrefmark{3}author.three@add.on.net,
% \IEEEauthorrefmark{4}author.four@add.on.net}}

\author{Haoran Wang \thanks{Three authors contribute equally to this work.} \\
Xidian University \\
{\tt\small wanghaoran@stu.xidian.edu.cn}
\and
Yue Fan \samethanks\\
Max Planck Institute for Informatics\\
Saarland Informatics Campus\\
{\tt\small yfan@mpi-inf.mpg.de}
\and
Zexin Wang \samethanks\\
Xidian University\\
{\tt\small zexinwang2016@gmail.com}
\and
Licheng Jiao\\
Xidian University\\
{\tt\small lchjiao@mail.xidian.edu.cn}
\and
Bernt Schiele\\
Max Planck Institute for Informatics\\
Saarland Informatics Campus\\
{\tt\small schiele@mpi-inf.mpg.de}
}

\maketitle
%\thispagestyle{empty}

\begin{abstract}
%\bernt{In my view we should call the spatial attention `parameter-free' and not `non-parametric'. In density estimation non-parametric is something quite different than what we have here}
%Global average pooling has a strong localization ability during the recognition \cite{CAM}. It helps the convolution neural network pay attention to the most discriminative feature on the image, thus vulnerable if the feature is missing due to the change of the camera viewpoint. We show that this is because of the lack of the spatial relation among the high-level features, which is advantageous in helping the model attend to the whole configuration of the object. We thus propose a non-parametric spatial attention layer which introduces the spatial relation among the activations on the last feature map back to the model. Based on that, we propose a novel architecture for Person Re-Identification task. Despite easy implementation, it consistently improves the performance over the model without it. The evaluation on 3 benchmarks demonstrates a superiority of spatial attention layer to all the state-of-the-art methods. Our model achieves rank-1 accuracy of 95.0\% on Market-1501, 89.0\% on DukeMTMC-ReID and 74.9\% on CUHK03-labeled. The corresponding mAPs are 91.9\%, 85.9\% and 76.5\%.

Global average pooling (GAP) allows to localize discriminative information for recognition \cite{CAM}. While GAP helps the convolution neural network to attend to the most discriminative features of an object, it may suffer if that information is missing e.g. due to camera viewpoint changes.
%thus vulnerable if the feature is missing due to the change of the camera viewpoint.
To circumvent this issue, we argue that it is advantageous to attend to the global configuration of the object by modeling spatial relations among high-level features.
We propose a novel architecture for Person Re-Identification, based on a novel parameter-free spatial attention layer introducing spatial relations among the feature map activations back to the model.
%Based on the non-parametric spatial attention layer
Our spatial attention layer consistently improves the performance over the model without it. Results on four benchmarks demonstrate a superiority of our model over the state-of-the-art achieving rank-1 accuracy of 94.7\% on Market-1501, 89.0\% on DukeMTMC-ReID, 74.9\% on CUHK03-labeled and 69.7\% on CUHK03-detected. % The corresponding mAPs are 91.9\%, 85.9\% and 76.5\%.

%On the other hand, it also leads to a fast convergence for the model since it helps stabilize the gradient flow.
%This paper proposes a non-parametric spatial attention layer for neural networks as a new approach to Person Re-Identification (Re-ID) task. It introduces the spatial relation among the activations on the last feature map back to the model, which is neglected by Global Average Pooling (GAP). Spatial relation among the high-level features is advantageous in helping the model attend to the configuration of body parts if the receptive field doesn't cover the whole image. On the other hand, it also leads to a fast convergence for the model since it helps stabilize the gradient flow. Despite easy implementation, non-parametric spatial attention layer consistently improves the performance on Person Re-ID task without the overhead cost. Evaluation on 3 benchmarks demonstrates a superiority to all the state-of-the-art methods. Our model achieves rank-1 accuracy of 95.0\% on Market-1501, 89.0\% on DukeMTMC-ReID and 74.9\% on CUHK03-labeled. The corresponding mAPs are 91.9\%, 85.9\% and 76.5\%.

%Improvement of the deep supervision for a better balance among losses.

%A performance gain by xxx is also observed on CUHK03-NP which is known for the scarce data.
\end{abstract}

\section{Introduction}
% Recently, this field has already seen a lot of great improvements with the incorporation of techniques from deep neural network.
The aim of Person Re-Identification (Re-ID) is to match people from different camera viewpoints. The task is challenging because of  significant pose-variations, frequent occlusions, and different camera viewpoints.

\begin{figure}[t]
  \begin{center}
  \includegraphics[width=0.9\linewidth]{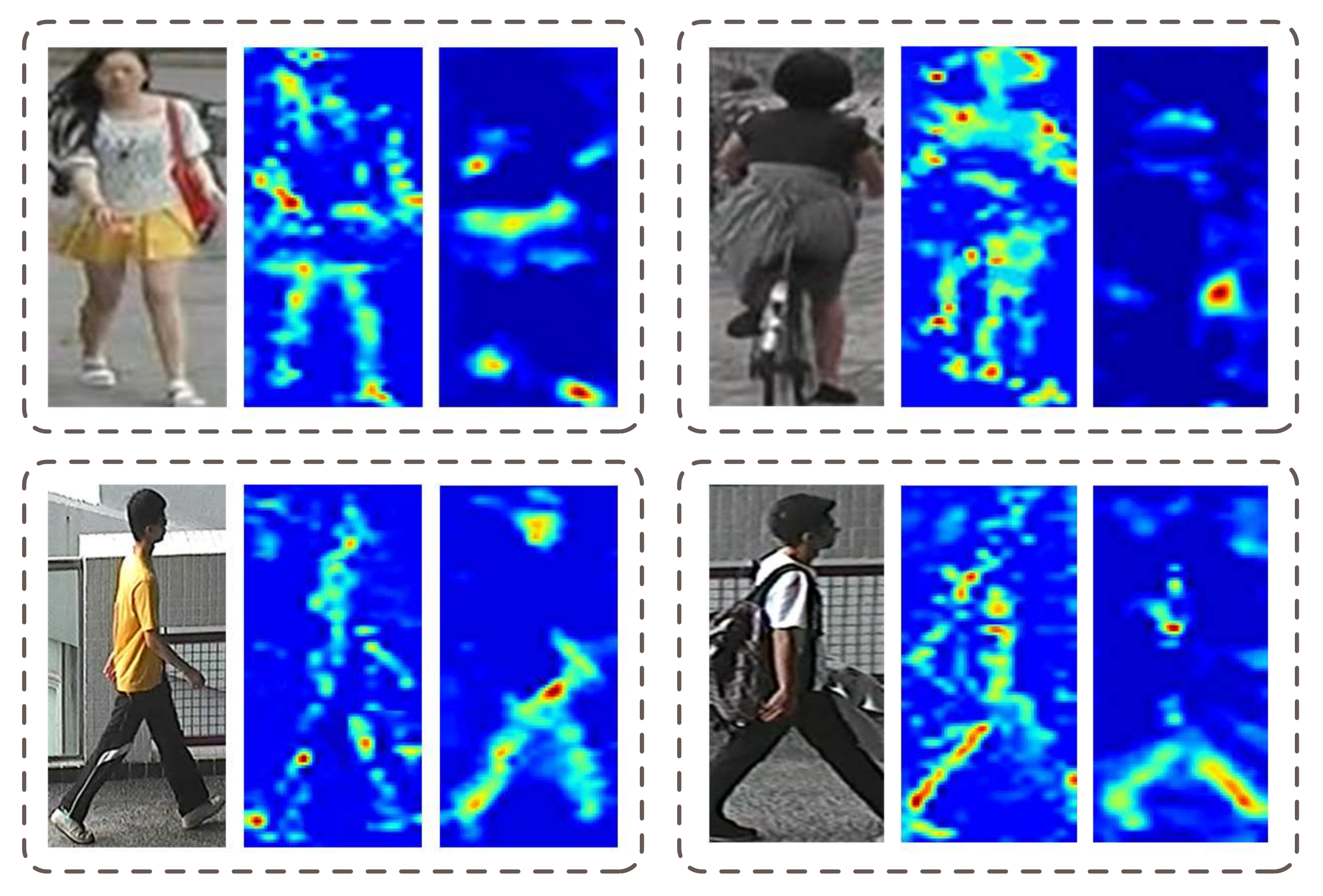}
  \end{center}
  \caption{Class activation maps (CAM) for four training images. In each example, from left to right are the original image, the CAM from GAP with our proposed parameter-free spatial attention and the CAM from plain GAP. The highlighted area is less concentrated with the help of the spatial attention.}%\bernt{the example at the bottom right is not as convincing as the other three to me visually}
  \label{fig:C}
\end{figure}

Global Average Pooling (GAP) \cite{GAP} is a well-known technique to reduce the number of parameters in the fully-connected layer. GAP often improves model performance due to a regularization effect of the capacity.
%It has a wide range of application in classification and detection \cite{GAPforc, GAPford}.
Zhou \etal \cite{CAM} analyzed the ability of the GAP layer to localize important regions from the input image even when the convolutional neural network is merely trained on image-level labels. This localization can be thought of as an implicit attention mechanism. At the same time, they proposed the class activation map (CAM) as a generic visualization method to show which regions of the image the model attends to when making a prediction. The third image in each example in figure \ref{fig:C} shows some class activation maps using plain GAP when the corresponding person is recognized. We can see that the highlighted area from GAP is always spatially concentrated. If the model fails to put the concentration onto the most discriminative region or if this region is missing in another camera view, then it may fail to identify that person. For example, GAP focuses on the waist of the first person in figure \ref{fig:C}, and if the person turns around in another image, the model might fail to match that person. This issue is less pronounced for general image classification as the classes are relatively distinctive. Therefore, the degradation from the absence of certain features is not necessarily fatal. But the Person Re-ID task is rather a fine-grained classification task, where the combination of many details and feature patterns of the object matter. Thus, the model should focus on the overall feature patterns and their relations. We will show that the spatially concentrated attention of GAP is  due to the lack of the spatial relation among the activations on the feature map in section \ref{ICAM} and we thus propose a new model to re-introduce spatial relations.
%Thus, the spatial relation is missing here.

%Among all approaches that have been proposed to address the issue of spatial relation among activations,
Attention-based methods are particularly promising to model spatial relations~\cite{atten_reid_harm, atten_reid_harm1, atten_reid_harm2, atten_reid_harm3, atten_reid_harm4, atten_reid_harm5}. Attention mechanisms \cite{attention_origin} equip the model with the ability by which it can pay attention to the most informative regions rather than the entire image during  recognition. Inspired by these mechanisms, we proposes a parameter-free spatial attention layer to incorporate the spatial relation for GAP. The second image in each example in figure \ref{fig:C} shows the change of the class activation maps when our spatial attention layer is employed. From the images we can observe that the attention is, on the one hand, more distributed over the image than for GAP, but, on the other hand, also attends to regions that are attended to by GAP. Such attention thus can be more robust to partial occlusions or camera viewpoint changes.

\begin{figure}[!ht]
  \centering
  \includegraphics[width=0.6\linewidth]{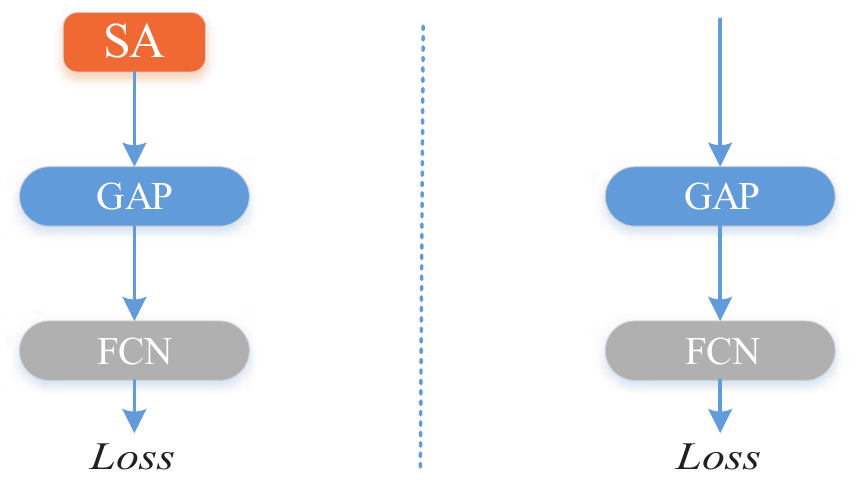}
  \caption{The structure of a common classifier with GAP is shown on the right; the modification on the left is that a spatial attention layer (SA) is inserted before GAP.}
  \label{fig:overall_sa_gap}
\end{figure}

Figure \ref{fig:overall_sa_gap} shows where the spatial attention layer (SA) is added for a common GAP-based classifier. Instead of passing the feature map from the last convolution layer through GAP before a fully-connected layer, we first assign different \textit{importance} to different spatial positions with the help of the spatial attention layer. The importance of a spatial position is computed based on the total intensity of the activations along the corresponding channel. Therefore, it regularizes the GAP in a complementary way where all learned filters compete with each other to become more important. This leads to a distributed attention, thus covering more details of the image. We will show in section \ref{ICAM} that this is achieved by introducing the spatial relation on the feature map. From another perspective, the importance of an activation of a GAP-based model is only determined by its own intensity. In our case, however, it is also enhanced by other activations along the same channel, which in turn stabilizes the gradient flow, discussed in section \ref{BPSA}.

Another advantage of the spatial attention layer over the previous attention-based methods is that this module is entirely parameter-free. The increase of the calculation is only proportional to the number of times it appears in a model, which means it doesn't grow with respect to the scale of the dataset or the model size. Despite its simple structure, we show  experimentally in  section \ref{ablation} that it improves the performance consistently over the model without it.

%Global Average Pooling (GAP) \cite{NIN} is proven to be very useful in a lot of different settings. It makes the kernels to be competitive to each other so there will be no identical kernel. With GAP, you ignore the spatial relation on the last feature map, it can be seen as paying attention to one region of the image, this is good for xxx but it is bad for xxxx. Our SA saves the world by distributing the attention over the whole image. This is very important for Person-ReID, because of occlusion and camera viewpoint change. We can actually see that, GAP tends to look at the face. But if the guy turns around, then you are not robust any more. But we are robust(Thus more robust to occlusion). Also, this SA can be seen as a complementary module to GAP because it has no parameter except a bit more calculation during the training. Attention layer in normally trainable, for example in xx and xx. But ours is much cheaper. And further. We show the understanding part, that's why our SA works. Also we show that it can stabilize the gradient flow, which is awesome.

%In \cite{CAM}, Zhou et al. used a heat map to illustrate the contribution distribution from different image regions when a certain class of object is recognized.

Overall, our contribution is threefold. 1) We introduce a parameter-free spatial attention layer which shows consistent improvement for GAP. 2) We propose a novel architecture with the spatial attention layer for Person Re-ID and achieve state-of-the-art performance on four benchmarks. 3) We compare the difference of GAP with and without spatial attention analytically for a better understanding.
%\bernt{in the abstract we mention 3 benchmarks - not four?}
%Zhou et al. \cite{CAM} first shed light on that GAP layer are able to localize important regions from the input image even trained on image-level labels for convolutional neural networks, which can be thought of as an implicit attention mechanism. At the same time, they proposed a visualization method called class activation map to show which regions of the image the model attends to when making a prediction. We adopt the same method to visualize the impact on the class activation map for non-parametric spatial attention layer. Figure \ref{fig:C} shows the attention distribution (yet not a probability distribution) over the whole image when the person in the row above is recognized. The brighter regions implies more contribution to the decision. We can see from the examples that a lot of different regions are taken into account for the identification. Regions with high contrast or complicated patterns tend to abstract more attention, which is also the case for human beings. We compare a lot of different discriminative regions of the image when we identify an object.

%\cite{atten_reid_harm} and \cite{atten_reid_harm2} also demonstrated the effectiveness of the attention mechanism.

\section{Related Work}
%Most existing person re-id methods  first extract discriminative hand-crafted and deep learning features that are robust to illumination and viewpoint changes \bernt{please use properb citation keys - otherwise this will be a mess eventually} [9,13,24,32,40,41,59], and then use metric learning \bernt{citations keys please} [2,6,12,18,22,31,32,33,36,43,54,58,63] to ensure that features from the same person are close to each other while from different people are far away in the embedding space.
\myparagraph{Weakly Supervised Learning.}
There are many weakly supervised learning approaches using various pooling operations to localize objects, thus making use of more details on the feature map for better classification \cite{mantra, parizi2014automatic, MIL, isthere}.

Zhou \etal \cite{CAM} first showed the ability of GAP to localize the most discriminative image region. Instead of GAP, WELDON \cite{weldon} provides a more robust and automatic activation selection strategy for pooling by selecting multiple high and low score regions from the last feature map, which is a generalization of the min + max prediction function in \cite{mantra}. However, they just selected top positive and negative instances with the highest and lowest activations and then aggregated them. Different from them, we use all of the activations on the last feature map while having a fixed budget for the total importance by the softmax function. Note, that both are parameter-free methods.

\cite{wildcat} also adopted a similar idea for spatial pooling by introducing an extra hyper-parameter to trade off relative importance between positive and negative instances. But still, it doesn't take all activations into consideration. In our case, the importance of an activation is determined by its relative magnitude to others.

Based on the fact that lower convolution layers learn redundant filters to extract both positive and negative information due to the positive output from ReLU, Shang \etal \cite{crelu} and Blot \etal \cite{minmaxconv} proposed a Concatenated Rectified Linear Unit to preserve both positive and negative information by concatenating a negated copy of the same feature map before ReLU, thus preventing the model from learning highly negatively-correlated pairs of the filters. This has a similar effect as our spatial attention layer, which enforces the filters to compete with each other by setting a budget on the total importance with the help of the softmax function.

\begin{figure*}[!ht]
  \centering
  \includegraphics[width=0.9\textwidth]{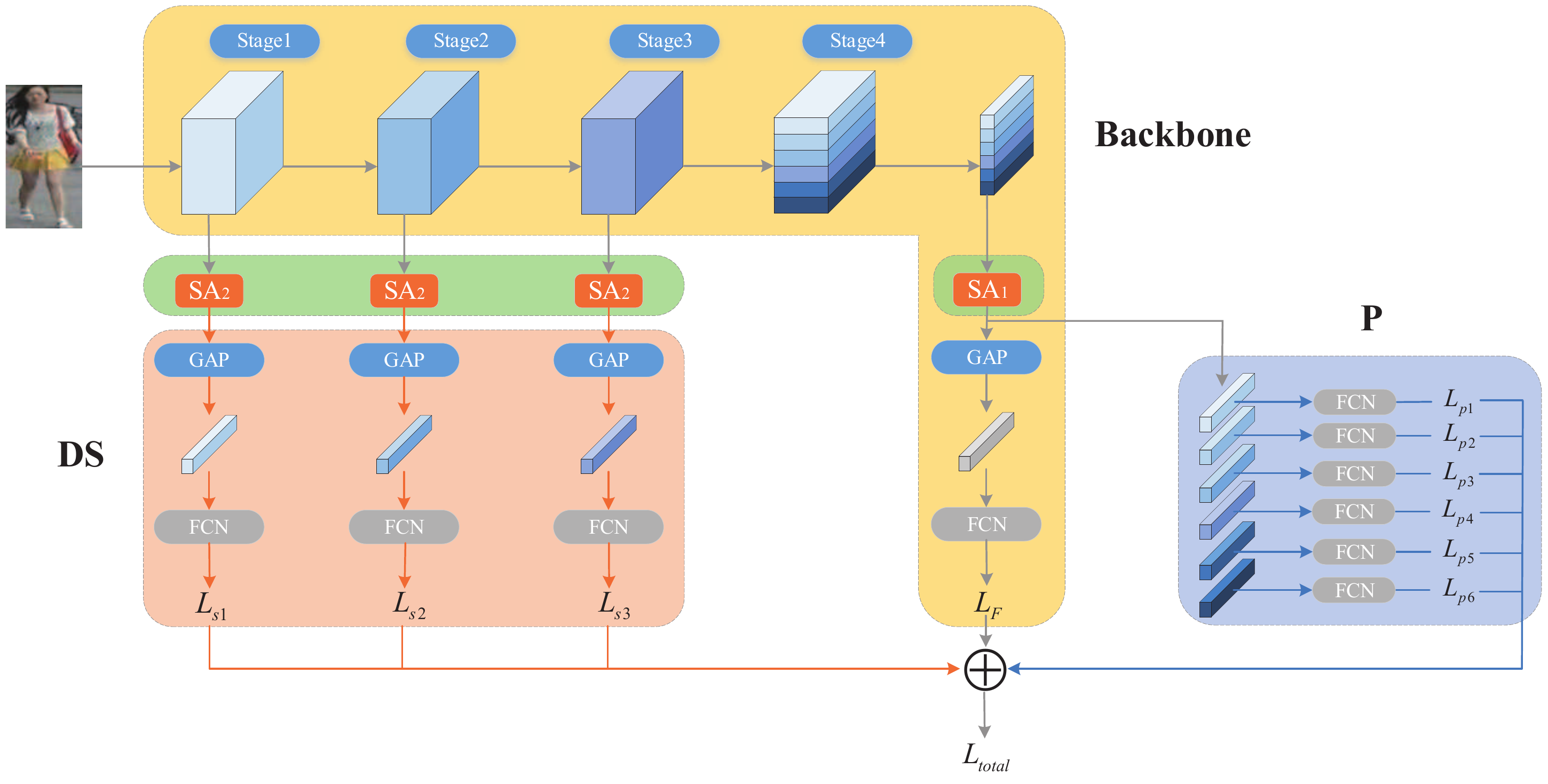}
  \caption{The proposed architecture formulates the task as a classification. It consists of four components. The yellow region represents the backbone feature extractor. The red region represents the deeply supervised branches (DS). The blue region represents six part classifiers (P) \cite{pcb}. The two green region represents two sets of spatial attention layers (SA), SA1 is not used for the main results. It only appears in the section \ref{ablation} for the sake of ablation study. Then the total loss is the summation over all deep supervision losses, six part losses and the loss from the backbone. Note that the spatial attention is only added before GAP.}
  \label{fig:train_arch}
\end{figure*}

\myparagraph{Attention in Person Re-ID.}
The attention mechanism proposed in \cite{attention_origin} has shown great success for a broad range of computer vision tasks. There also exists many attention-based approaches for Person Re-ID.

One way to introduce attention is to insert a trainable layer into the main body of the model \cite{atten_reid_harm}. However, it allows the model to attend not only to the different spatial locations but also to the different channels with various weight, which means the model has to decide for every single activation inside a feature map whether it is worth to pay attention to it. This, on the one hand, gives the model a lot of flexibility to make use of the most valuable features, but on the other hand can be harmful if no proper regularization techniques is applied to reduce overfitting. Another similar work is from Hu \etal \cite{senet}, where the feature map was first squeezed by GAP and then fed through two fully-connected layers before rescaled by a sigmoid function. The resulting vector had the same length as the number of channels of the original feature map, onto which each entry of the vector was multiplied back for the corresponding channel. It differs from our work in three aspects: we use softmax rather than sigmoid for the resulting vector; the attention in their work was mainly spread among the channels rather than the spatial positions; and our parameter-free module is only used before the last GAP layer.

\myparagraph{Deep Supervision and Part-Level Feature.}
The idea of deep supervision was first proposed by Lee \etal in \cite{deepsupervision}, where a progressively decay for the weight was adopted on the auxiliary losses which went to zero eventually. The auxiliary losses can lead to more discriminative intermediate features.
%The portion between the main loss and the auxiliary losses was determined by a parameter $\alpha$ which decays with the increase of the epoch number $t$:
%$$\alpha_t = \alpha_0 (1 - \frac{t}{N})$$
%where $\alpha_0$ is the initial weight and $N$ is the total number of epochs.
This technique was introduced by Wang \etal \cite{dare} for the first time to the Person Re-ID task. They took the summation over all the losses as their final loss, which implicitly assigned equal weight to each individual loss. In this work, we use a fixed proportion for the main and auxiliary losses, namely, $0.8$ and $0.2$. It achieves empirically similar results as the decay strategy in \cite{deepsupervision}.
%For the sake of comparison, we further train two models adopting the strategies from \cite{dare} and \cite{deepsupervision} respectively. \cite{dare} put the same weights on both types of loss, which leads to 92.3\% R-1 and 77.7\% mAP. \cite{deepsupervision} used $\sum_{p=1}^{m}{l_p} + \alpha_t \sum_{i=1}^{3}{l_{s_i}}$ as the final loss $L$. In our experiment, $\alpha_0 = 0.3$ and $N = 150$, the corresponding top 1 is xxx and mAP is xxxx. Therefore, a fixed portion of weights on different losses is more beneficial in our task. The reason for the inferiority of equally weighting strategy might lie in the fact that the heavy regularization consistently contributes gradients back to the main stream even when the model is approaching a local minimum. This extra flow makes the model harder to converge thus degrading the performance.

%We add different weights to the deep supervised losses to balance the focus of the training so that more attention should be payed on obtaining more discriminative high-level feature.
Part-level features have shown powerful performance in Person Re-ID due to the robustness to partial occlusion and camera view changes \cite{pcb, pcb1, pcb2, pcb3}. Sun \etal showed a strong baseline model in \cite{pcb}, where the last feature map was uniformly partitioned into six horizontal stripes before being averaged into a single-channel part-level feature. Then six independent losses were applied for them, respectively. The final loss was simply the summation of all six losses. We leverage this technique along with a seventh global loss which is obtained by taking the six horizontal stripes as a whole.
%This minor modification actually shows a quite different viewpoint. In our formulation, each part-level descriptor contains only partial semantic information from the image. Therefore, using it for classification will basically impose a strong drop out affect to the model. The model should be able to find the correct person given, for example, only the upper part of the body.

%Recently, [Person re-identification with correspondence structure learning][ Partial person re-identification], both patches-matching and stripes-matching methods [Person re-identification with correspondence structure learning][ Partial person re-identification] [Constrained deep metric learning for person re-identification] are proved to be effective. Considering that in the deeper network layer, the feature field of each pixel level is sufficiently large in the original image, We use patch-level features and strips-level features at the shallower network layer and the deeper network layer, respectively, Which brings a pretty significant improvement.

\section{Proposed Method}

% \begin{figure}[!ht]
%   \centering
%   \includegraphics[width=0.5\textwidth]{img/SA.eps}
%   \caption{Non-parametric Spatial Attention Layer has no trainable parameter. It assigns different importance for different locations. It consists of a summation and a softmax. $A(i,j)$ denotes the summation over the activations along the channel. The softmax values are computed over $A$ then multiplied to all activations from the same location.}
%   \label{fig:SA}
% \end{figure}

This section will propose a novel architecture for Person Re-ID that is, on the one hand, a novel combination of power-full components proposed in the literature, and, on the other hand, also contains our novel parameter-free spatial attention layer resulting in consistent improvements.

%In order to apply the idea of attention, we first sum over channels for each spatial position, we get A, which is an overall condition. To make it competitive, we use softmax for it rather than sigmoid. in this case, these activation will start to fight with each other to draw more attention. This competition actually has a regularization affect, which makes each channel more distinctive, they learn different things. So for example, if one kernel has already xxxxx. So the importance of each spatial position is determined by the summation of activations along the channel.

\subsection{Network Architecture} \label{net}

\begin{figure*}[!ht]
  \centering
  \includegraphics[width=0.85\textwidth]{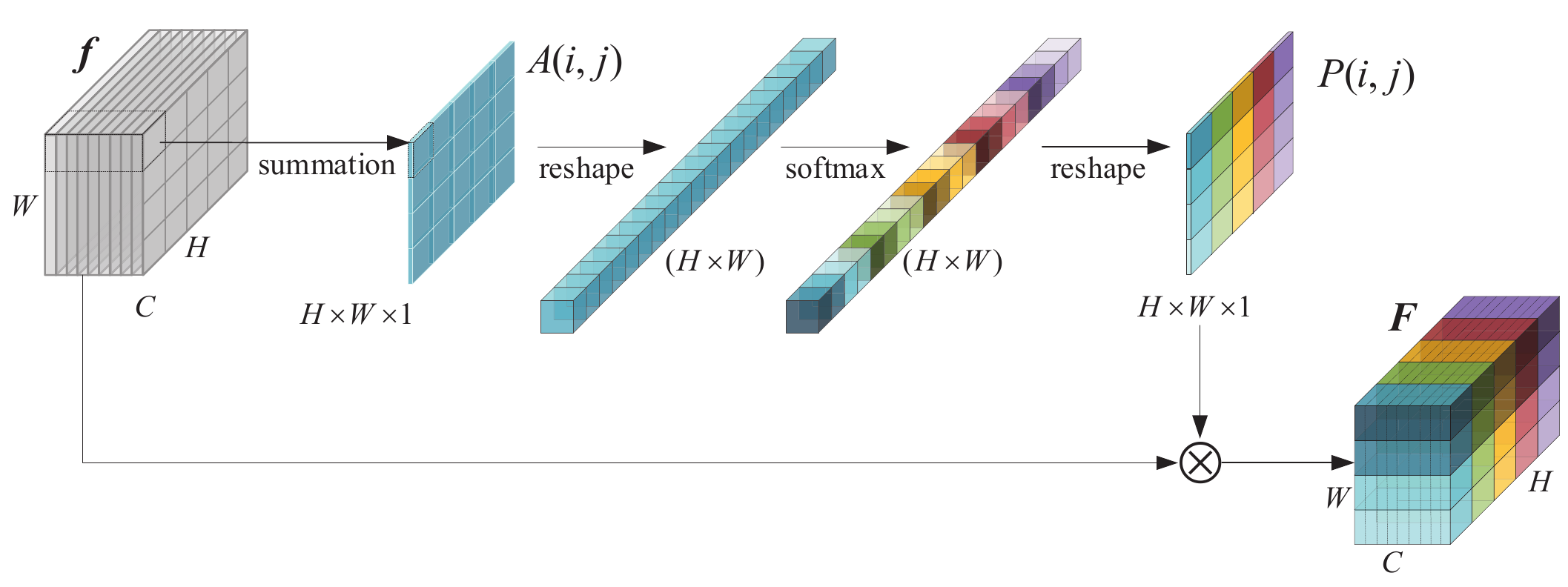}
  \caption{Parameter-free spatial attention layer has no trainable parameter. It assigns different importance for different locations. It consists of a summation and a softmax. $A(i,j)$ denotes the summation over the activations along the channel. The softmax values are computed over $A$ then multiplied to all activations from the same location.}
  \label{fig:SA}
\end{figure*}

%Figure \ref{fig:train_arch} shows the overall architecture. The blue part is the backbone, the green part is deep supervision, the red part is pcb. Our SA is added at two different locations, deep supervision branch and the main branch. It is illustrated in the figure 3.

%Besides the main pipeline mentioned above, three green branches are also shown in figure \ref{fig:train_arch}. Each branch represents a deep supervision with the help of the non-parametric attention layer.

We formulate the Person Re-ID task as a classification problem.
Our overall architecture leverages several established and powerful techniques and is enhanced by our novel spatial attention layer.
In particular we leverage the following techniques:
we adopt a standard ResNet-50 up to stage 4 for the backbone as it has shown good performance for our task \cite{pcb, dare, aacn};
we also employ deep supervision due to its ability to also influence the lower feature layers of our network \cite{dare}; and we also partition the last layer according to PCB \cite{pcb} which uniformly partitioned the last feature map from the backbone in order to produce a loss for each part-level feature. PCB is proven to give competitive results in Person Re-ID task, too.
Therefore, combining the ideas from \cite{dare} and \cite{pcb}, we propose a novel architecture as shown in the figure \ref{fig:train_arch}. The yellow part is the backbone ResNet-50. The red part represents the deep supervision branches, for which the spatial attention layers (the left green part) are added before GAP. The blue part denotes the part classifiers from \cite{pcb}. Each part classifier produces a loss based on the part-level feature. Note that the right spatial attention layer in green is not a part of the main model. It is only used in the ablation study in the section \ref{ablation}.
%useful in this task for a lot of times \bernt{please give references here}. Deep supervision is also nice, we use it from Dare \bernt{please give reference here}, partitioning the last layer is also very nice according to pcb \bernt{please give reference here}. We combine all these good things together. And also we add the spatial attention. it gives consistent improvement in section 5.2.

%Since ResNet \cite{resnet50} has been shown to have superior classification performance, we adopt a standard ResNet-50 up to stage 4 as the backbone. \cite{dare} shows the effectiveness of deep supervision in Person Re-ID task. \cite{pcb} introduces a uniform partition strategy to process the last feature map from the backbone. It produces a loss for each partitioned feature vector. The final loss is simply the summation of all the losses. This method is proven to give competitive results in Person Re-ID task, too.

The input for every deeply supervised branch is a feature map $f$. The spatial attention layer first assigns different importance for different locations. The processed feature map is then fed into a GAP layer followed by a fully-connected layer before computing the cross-entropy loss $l_{s_i}$. The losses calculated from three intermediate feature maps will in the end be added to the final loss. Therefore, our final loss $L_{total}$ consists of three auxiliary losses, partition losses and the main loss, that is $L_{total} = (1 - \lambda) \sum_{p=1}^{m}{l_p} + \lambda \sum_{i=1}^{3}{l_{s_i}} + \lambda l_F$, where $\lambda$ controls the proportion of losses. The $L_F$ uses the same weight with the auxiliary losses because the backbone can be also thought of as another deeply supervised branch. The only difference is that the feature vectors obtained from stage 4 will be used in inference.

During training, the yellow backbone first takes a image as the input, then the extracted feature map whose size is $H \times W \times C$ at the last layer is then vertically partitioned into $m$ segments, over each of which GAP is then performed. The resulting part-level feature is of size $1 \times 1 \times C$. Afterwards, each part-level feature is convolved with a shared $1 \times 1$ filter into the shape $1 \times 1 \times \Hat{C}$ before an independent linear classifier with cross-entropy loss is deployed. Up to now, $m$ different \textit{partition losses} $l_p$ are computed. Taking all $m$ reshaped part-level features as a whole, another GAP is applied to obtain a $1 \times 1 \times \Hat{C}$ feature which is then used to compute the main loss $L_F$. The partition loss should be thought of as a dropout \cite{dropout} for high-level features. It has a significant regularization effect on the model because each partition loss is independent of each other, thus forcing each part-level feature to be discriminative enough to yield the correct prediction. This is more efficient than random erasing \cite{random_erase} since the latter one is applied on the pixel level.

During inference, all the deeply supervised branches and part classifiers on top are dismissed. The backbone will then be used as a feature extractor, which is applied to all the query images and the images in gallery to obtain the features. Then a standard information retrieval is performed based on the Euclidean distance between these features.

\subsection{Parameter-Free Spatial Attention Layer} \label{31}

The spatial relation among the activations is helpful to distribute model attention. However, GAP treats the activations on the same feature map equally despite their locations, making the model less robust to the absence of certain features. Our spatial attention layer can add this information back into the model.
Figure \ref{fig:SA} shows the details. The first operation is a summation over the channels for each spatial position, which indicates the importance of that position. To make the learned filter more competitive, softmax rather than sigmoid is then applied on the summed activations. This encourages different filters to learn different features, thus making the model more robust. The output from softmax is then used to rescale the magnitude of the activations on the original feature map. This implies that the importance of different activations at the same spatial position can be enhanced by other activations along that channel. All activations from the same spatial position are rescaled by the same weight while different spatial positions have different weights. The spatial relation is considered because the weight is proportional to the relative magnitude of the summed activations at the corresponding position.

%As detailed in the figure \ref{fig:SA}. The input for this layer is a feature map $f$, and the output is another feature map $F$ with the same shape.

Formally, given a feature map $f$ with size $H \times W \times C$, a summation along the channel axis yields a 2-D matrix $A$ with size $H \times W$, where $A(i, j) = \sum_{k = 0}^{C}{f_k(i, j)}$. Then a softmax function is applied to the flattened matrix $A$ in order to assign each spatial position a value $p(i, j)$ indicating the degree of importance for that location. The afore-produced values will be multiplied to all the activations along the channel axis of $f$ for the corresponding spatial positions. Therefore, the output $F$ of a parameter-free spatial attention layer can be written as:
\begin{equation}
    F_k(i, j) = f_k(i, j) p(i, j) \label{eq:0} \tag{1}
\end{equation}
where $p(i, j) = \frac{e^{A(i, j)}}{\sum_{i,j}{e^{A(i, j)}}}$.

%The non-parametric attention layer assigns different importance degrees for different locations. Previously, \cite{dare} simply added all \textit{per-stage loss} onto the final loss without weighing, which indicates they are of the same importance implicitly. This can be sometimes harmful since these per-stage losses are just supplementary to the main loss in the hope that more robust high-level feature can be extracted from more discriminative low-level features. In the end of day, it is, however, still the high-level features that are involved in the inference phase. Following a general machine learning scenario, per-stage loss should rather play a role of regularization which needs a parameter $\lambda$ to make control the focus of the training. Non-parametric attention layer is applied to 3 intermediate feature maps in the model above.

\subsection{Impact on the Class Activation Map} \label{ICAM}
The class activation map (CAM) proposed in \cite{CAM} visualizes the attention distribution over the input image when the model identifies a specific class. As shown in figure \ref{fig:C}, the spatial attention layer helps to distribute the focus of the model to other regions besides the most discriminative one. The class activation map of GAP of the first person highlights the waist and the feet of that person. But with spatial attention, other parts of the body which are also informative are highlighted as well.

Following the notations from section \ref{31}, $f$ is the feature map from the last convolution layer. If a fully-connected layer is directly performed on $f$ for classification, the logits $S_c$ for a specific class $c$ is then given as:
\begin{equation}
    S_c = \sum_{i, j, k}{w_{ijk}^c \cdot f_k(i, j)} = \sum_{i,j}{\sum_{k}{w_{ijk}^c \cdot f_k(i, j)}} \label{eq:1} \tag{2}
\end{equation}
where $w_{ijk}^c$ is the weight connecting the logits $S_c$ and $f_k(i,j)$. Therefore, the class activation map $M_c$ can be defined as:
\begin{equation}
    M_c(i, j) = \sum_{k}{w_{ijk}^c \cdot f_k(i, j)} \label{eq:2} \tag{3}
\end{equation}
If Global Average Pooling is employed, then the result $G_k$ after applying GAP can be computed as $\sum_{i,j}{f_k(i, j)}$. The corresponding logits for class $c$ is then:
\begin{equation}
    S_c = \sum_{k}{w_k^c G_k} = \sum_{i,j,k}{w_k^c \cdot f_k(i, j)} = \sum_{i,j}{\sum_{k}{w_k^c \cdot f_k(i, j)}} \label{eq:3} \tag{4}
\end{equation}
This leads to the following form of the class activation map:
\begin{equation}
    \hat{M_c}(i, j) = \sum_{k}{w_k^c \cdot f_k(i, j)} \label{eq:4} \tag{5}
\end{equation}
The advantage of GAP lies in its regularization effect. The parameters $w_{ijk}^c$ in the fully-connected layer become $w_k^c$ after adding the GAP layer, which means GAP operations reduce the model capacity by enforcing $w_{ijk}^c$ to be the same for any spatial position $i$ and $j$, thus preventing overfitting of the model. But from another aspect, it can also be harmful if  the spatial relation across the high-level features is completely ignored, which resembles the spirit of bag-of-words methods \cite{BOW}, thus suffering from the same drawback. Ideally, we want to make a compromise between these two extreme cases, where GAP neglects the spatial relation while directly performing fully-connected layer lacks effective restrictions. To this end, the parameter-free spatial attention layer shows a combination of advantages from both sides.

The formula for the logits $S_c$ with the parameter-free spatial attention layer can be written as:
\begin{equation}
    S_c = \sum_{k}{w_k^c G_k} = \sum_{i,j,k}{w_k^c \cdot F_k(i, j)} = \sum_{i,j}{\sum_{k}{w_k^c \cdot F_k(i, j)}} \label{eq:5} \tag{6}
\end{equation}
where $F$ is the output from the parameter-free spatial attention layer. Therefore, the class activation map will be:
\begin{equation}
    \Tilde{M_c}(i, j) = \sum_{k}{w_k^c \cdot F_k(i, j)} = \sum_{k}{w_k^c \cdot f_k(i, j) \cdot p(i,j)} \label{eq:6} \tag{7}
\end{equation}
It can be thought of as a factorization of $w_{ijk}^c = w_k^c \cdot p(i,j)$ where $w_k^c$ controls the inter-channel attention and $p(i,j)$ controls the spatial attention. The assumption behind this is that the spatial and channel attention are independent to each other. This factorization decomposes the coupling between them and at the same time maintains the spatial term. The magnitude of $p(i,j)$ depends on the number of activations and the intensity of each activation across all channels. Typically, large $p(i,j)$ indicates a region of interest in the input image which contains lots of desired features. Thus more attention should be payed to it.

%In the comparison with GAP, we can see that the only difference is a layer. But the output shape is the same. The only commputation is a summation and a softmax and a multiplication. Also, at the same time, softmax can make attention more distributed as is shown in figure 1 compared to GAP. Thus more robust to occlusion. Moreover, why do we do the summation along the channel, not just take the softmax over all the activation?? This is of course a better way, but this increase the model capacity, but we want regularize it. Because the channel info has already been addressed by the weight in classifier. So the assumption here is that the importance of an activation $W$ can be decomposed into wij and wc. Why is this reasonable?? Because this on the one hand reduce the capacity, also didn't miss the ij info, but GAP missed it. The importance of the activation of GAP is only determined by it's own, but in our case, it is also enhanced by your peers.

%Containing no trainable parameters, this operation makes the deep supervision branch very cheap. The only parameters are from the fully connected layer.

\subsection{Backpropagation through Parameter-Free Spatial Attention Layer} \label{BPSA}

This section aims to justify  the parameter-free spatial attention layer in terms of the gradient flow.

Given the last feature map $f = Model(X;W)$, the common strategy is to compute $G = GAP(f)$, and then logits for a specific class is $S_c = Wc^T \cdot G$. Therefore, the gradient for $W$ in the model is given as:
\[
\frac{\partial S_c}{\partial W} = \frac{\partial S_c}{\partial G} \frac{\partial G}{\partial f} \frac{\partial f}{\partial W} \label{eq:7} \tag{8}
\]
When the parameter-free spatial attention layer (SA) is used, then the forward pass will become $F=SA(f)$, $G=GAP(F)$, $S_c = Wc^T \cdot G$. And the corresponding backward pass will be:
\[
\frac{\partial S_c}{\partial W} = \frac{\partial S_c}{\partial G} \frac{\partial G}{\partial F} \frac{\partial F}{\partial f} \frac{\partial f}{\partial W} \label{eq:8} \tag{9}
\]
The only difference between \eqref{eq:7} and \eqref{eq:8} is an extra intermediate term $\frac{\partial F}{\partial f}$ that scales the gradients before it is passed further.

Following the notation defined in the section \ref{31}, we can compute the gradient of $F$ with respect to $f$ using the chain rule:
\[
\frac{\partial F_k(i,j)}{\partial f_t(m,n)} = f_k(i,j) \frac{\partial p(i,j)}{\partial f_t(m,n)} + p(i,j) \frac{\partial f_k(i,j)}{\partial f_t(m,n)} \label{eq:9} \tag{10}
\]
where the latter part doesn't vanish only if $\{k,i,j\} = \{t,m,n\}$.

The derivative of the softmax function is given as:
\[
\frac{\partial p(i,j)}{\partial a(m,n)} =
\begin{cases}
p(i,j)(1-p(i,j)), &\{i,j\} = \{m,n\}\\
-p(i,j)p(m,n), & otherwise
\end{cases}
\label{eq:10} \tag{11}
\]
Plugging in the above equations, we can conclude that
\begin{align*}
& \frac{\partial F_k(i,j)}{\partial f_t(m,n)} =
\end{align*}
\[
\begin{cases}
f_k(i,j)p(i,j)(1-p(i,j)) + p(i,j), \;\; \{k,i,j\} = \{t,m,n\}\\
f_k(i,j)p(i,j)(1-p(i,j)),\;\;\;\;\;\;\;\;\;\; k \neq  t \land \{i,j\} = \{m,n\} \\
-f_k(i,j)p(i,j)p(m,n),\;\;\;\;\;\;\;\;\;\;\;\;\;\; else
\end{cases}
\label{eq:11} \tag{12}
\]

The derivation shows that the gradient which flows backwards from a certain activation $f_k(i,j)$ will be effected by three different sources, that is, the activations at the same spatial position but different channel, itself and others. Since $p(i,j)$ and $(1-p(i,j)$ are always non-negative, $\frac{\partial F}{\partial f}$ for the first two cases has the same sign as $f_k(i,j)$, which means that the activations from other channels but the same spatial position boost the gradient flow according to $f_k(i,j)$. Other activations, on the contrary, always have the opposite sign to $f_k(i,j)$ indicating a cancel-out effect to the gradient. Therefore, it stabilizes the gradient in a competitive way so that the training procedure suffers less from the vanishing or exploding gradient problem, which leads to fast convergence.

\section{Experimental Setup} \label{exp}
%\bernt{a better name for this section would be in my view `Experimental Setup'}
%\bernt{personally i would again prefer myparagraph than subsection for the two partitions of this section - I change them now but feel free to change back if you think differently - I am also defining to mypagraphs for the section subsection}
%\subsection{Datasets and Protocols}
\myparagraph{Datasets and Protocols.}
The experiments are evaluated on four large-scale datasets Market-1501 \cite{market1501}, DukeMTMC-ReID \cite{duke1} \cite{duke2} and CUHK03-NP \cite{cuhk03}.

Market-1501 contains 32,668 annotated bounding boxes of 1,501 identities which are taken from six different camera views. The identities are split into 751 training IDs and 750 query IDs. There are in total 3,368 query images, each of which is randomly selected from each camera so that cross-camera search can be performed. The retrieval for each query image is conducted over a gallery of 19,732 images including 6,796 junk images. DukeMTMC-ReID dataset uses the the same format and evaluation protocols as Market-1501. It contains 16,522 training images of 702 identities, 2,228 query images of the other 702 identities and 17,661 gallery images from 8 high-resolution cameras.
CUHK03-NP is re-formulated from the old CUHK03 dataset. It is designed for the new training/testing protocol proposed by Market-1501 \cite{market1501}. It contains 767 identities with 7,368 images and 700 identities with 5,328 for training and testing respectively. Each identity is observed from two non-overlapping cameras. The difficulty is thus increased since there are less ground truth images for a single query.

The evaluation protocol proposed in \cite{market1501} is used for Market-1501 and DukeMTMC-ReID. The Cumulative Matching Characteristic (CMC) for rank-1, rank-5 and the mean average  precision (mAP) are measured. The new protocol from \cite{cuhk03} is employed for CUHK03-NP dataset. Note that all the following results are evaluated under the single-query mode.

%\subsection{Performance Evaluation}

\begin{table*}[]
\begin{center}
\begin{tabular}{c|cc|cc|cc|cc}
\hline
\hline
Datasets & \multicolumn{2}{c|}{Market-1501} & \multicolumn{2}{c|}{DukeMTMC-ReID} & \multicolumn{2}{c|}{CUHK03 labeled} & \multicolumn{2}{c}{CUHK03 detected} \\ \hline
Metrics(\%) & R-1 & \multicolumn{1}{c|}{mAP} & R-1 & mAP & R-1 & mAP & R-1 & mAP \\ \hline
SVD-net \cite{random_erase} & 89.1 & \multicolumn{1}{c|}{83.9} & 84.0 & 78.3 & - & - & 41.5 & 37.6 \\
OIM loss \cite{oimloss} & 82.1 & \multicolumn{1}{c|}{60.9} & 68.1 & 47.4 & - & - & - & - \\
DPFL \cite{dpfl} & 88.9 & \multicolumn{1}{c|}{73.1} & 79.2 & 60.6 & 43.0 & 40.5 & 40.7 & 37.0 \\
PAN \cite{pan} & 82.2 & \multicolumn{1}{c|}{63.3} & 71.6 & 51.6 & 36.9 & 35.0 & 36.3 & 34.0 \\
FMN \cite{fmn} & 87.9 & 80.6 & 79.5 & 72.8 & 46.0 & 47.5 & 47.5 & 48.5 \\ \hline
HPM \cite{hpm} & 94.2 & 82.7 & 86.6 & 74.3 & - & - & 63.9 & 57.5 \\
PCB+RPP \cite{pcb} & 93.8 & \multicolumn{1}{c|}{81.6} & 83.3 & 69.2 & - & - & 63.7 & 57.5 \\
HA-CNN \cite{atten_reid_harm} & 91.2 & \multicolumn{1}{c|}{75.7} & 80.5 & 63.8 & 44.4 & 41.0 & 41.7 & 38.6 \\
Mancs \cite{mancs} & 93.1 & \multicolumn{1}{c|}{82.3} & 84.9 & 71.8 & 69.0 & 63.9 & 65.5 & 60.5 \\
DaRe(R)+RE+RR \cite{deepsuper_reid} & 90.8 & \multicolumn{1}{c|}{85.9} & 84.4 & 79.6 & {\color[HTML]{333333} 72.9} & 73.7 & {\color[HTML]{3166FF} \textbf{69.8}} & 71.2 \\
\hline
\textbf{Ours} & {\color[HTML]{3166FF} \textbf{94.7}} & \multicolumn{1}{c|}{{\color[HTML]{3166FF} \textbf{91.7}}} & {\color[HTML]{3166FF} \textbf{89.0}} & {\color[HTML]{3166FF} \textbf{85.9}} & {\color[HTML]{3166FF} \textbf{74.9}} & {\color[HTML]{3166FF} \textbf{76.5}} & 69.7 & {\color[HTML]{3166FF} \textbf{72.2}} \\ \hline \hline
\end{tabular}
\end{center}
\caption{Comparison between our model and other state-of-the-art methods. The highest value for each category is marked in blue.}
\label{table:main}
\end{table*}

\myparagraph{Implementation Details.}%\bernt{I change my mind nad lef another comment above - I would call the entire section Experimental Setup}
We use a similar setting to \cite{pcb}. The backbone model is a pre-trained ResNet-50 from ImageNet. All three deeply supervised branches take the feature map from the corresponding convolution stage as  input. The spatial attention layer and the GAP layer are applied before the fully-connected layer. Dropout is not used in the deeply supervised branches.

The feature map after the fourth convolution stage is treated differently according to \cite{pcb}. It is first partitioned into six part-level features. Then average pooling is performed within each feature. The resulting feature map is then convolved by a $1 \times 1 $ filter to reduce the dimension of each feature vector from $2048$ to $256$ before being processed further. The six part-level features also produce six independent losses.

The input person images are all resized to $384 \times 128$. Random erasing \cite{random_erase} and random horizontal flipping with 0.5 are applied as the data augmentation. Re-ranking strategy from \cite{reranking} is also used to further improve the performance. We adopt the recommended number of partitioning from \cite{pcb}. Dropout rate for Market-1501 is 0.47 and 0.5 for other datasets.
Pre-trained weights from Imagenet are used. Batch size is set to 48. Stochastic Gradient Descent (SGD) with momentum is deployed as the optimizer. The base learning rate starts from 0.1 and decays to 0.01 after 40 epochs. The learning rate for all  pre-trained layers is set to 0.1 times the base learning rate. All  models are trained until convergence. The final loss consists of 0.8 partition losses and 0.2 deeply supervised losses.

%The dropout rate is tuned on different datasets with $0.4$ for Market-1501, $0.5$ for DukeMTMC-ReID and $0.5$ for CUHK03-NP. The final loss consists of $0.8$. What about the validation. We change the stride in the last layer as 1. We use 0.8 0.2 for the deep supervised loss and the main loss. In stage 3 deep supervision is not addressed. We use the re-ranking from []. 2048 feature after stage 4 is convolved into 256 together with a relu and BN. The based ResNet-50 network is pre-trained on ImageNet dataset.  We adopt random horizontal flipping and random erasing [?] as the data augmentation. The batch size and the training epoch is set to 48 and 100 respectively. At each iteration, we sample a mini-batch of samples in a random way to prevent the risk of overfitting.  The SGD optimizer is used with the initial learning rate 0.1, and multiplied by 0.1 per 40 epochs. The weight decay, and the momentum are set to 2 ¡Á 10?4, and 0.9, respectively.

\section{Experimental Results and Discussions}
%\bernt{this section should be called `Experimental Results' in my view}
%In this section, we test the necessity for spatial attention and the regularization affect of deep supervision. In order to investigate the behavior of different modules,
\subsection{Main Results}
Evaluated on the four aforementioned datasets, the proposed architecture achieves state-of-the-art performance on Market-1501, DukeMTMC-ReID and CUHK03-NP-labeled datasets regarding both rank-1 accuracy and mAP as shown in table \ref{table:main}. Especially on DukeMTMC-ReID, it shows a dramatic improvement of 3.3\% for R-1 and 6.3\% for mAP. For the CUHK03-NP-detected dataset, the model is slightly worse (0.1pp) than the best model but is still competitive and outperforms the third best method by a significant margin.
%achieves  0.1pp less than the best model b is still competitive.
But for the other datasets, our model achieves top performance by a large margin.

\subsection{Analysis of the Class Activation Map}

\begin{figure}[!ht]
  \centering
  \includegraphics[width=0.45\textwidth]{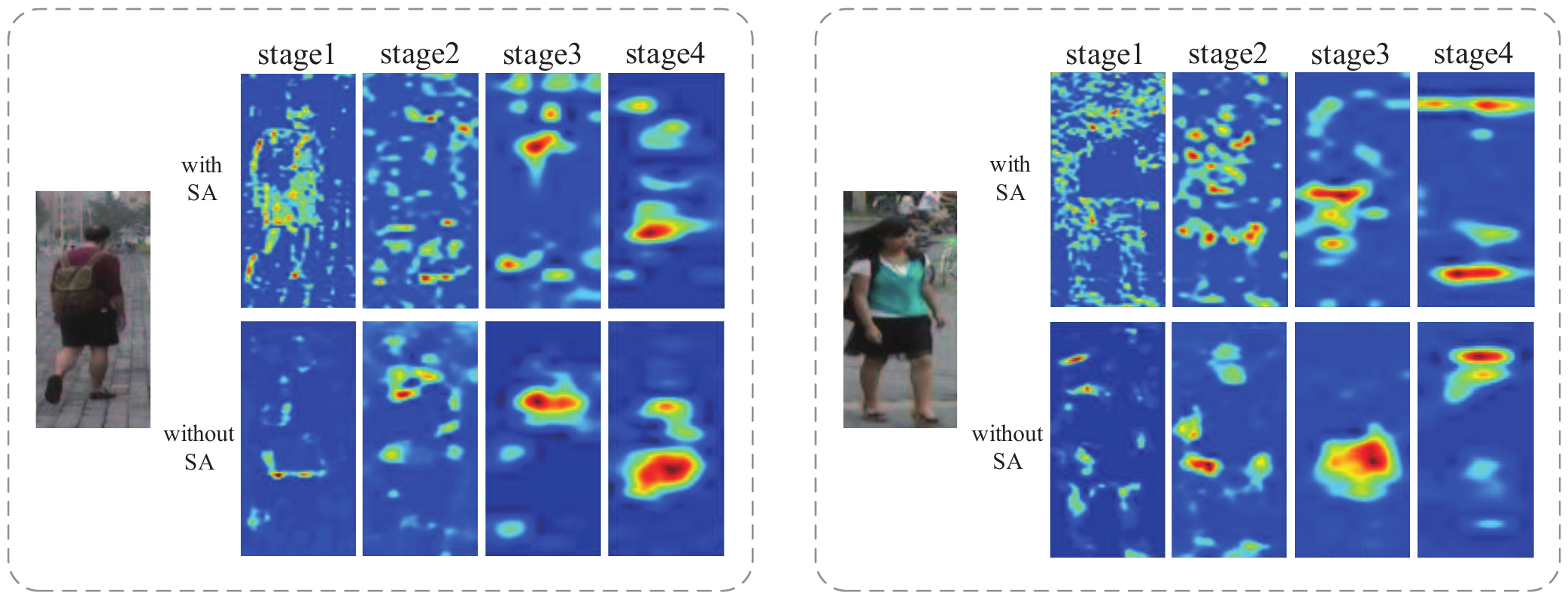}
  \caption{We visualize CAM of four stages for the model with and without the spatial attention. In both examples, the model with spatial attention shows a properly distributed focus for every stage.}
  \label{fig:stage_cam}
\end{figure}

\begin{figure*}[!ht]
  \centering
  \includegraphics[width=1.0\textwidth]{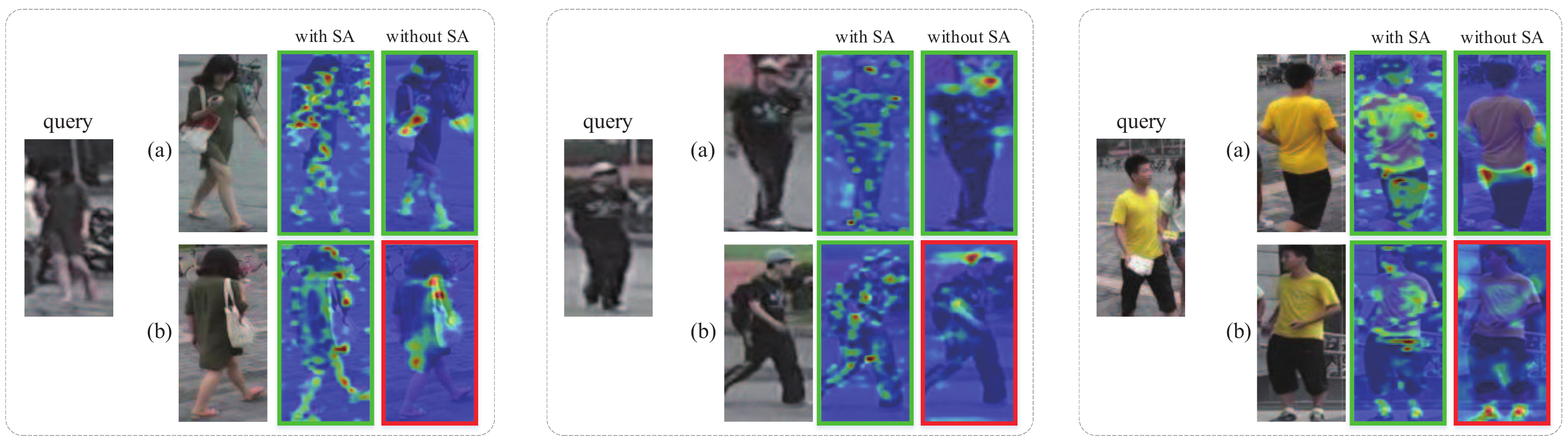}
  \caption{Given a query image and two test images, (a) is taken from the same camera view with the query and (b) is from another one. The third column of each example shows the original class activation maps. The second column shows the class activation maps with the spatial attention. The failure cases are marked with a red box. All are taken from the stage 2.}
  \label{fig:cam}
\end{figure*}
%This visualization is about the stage 2.
Zhou \etal \cite{CAM} used class activation maps (CAM) to illustrate the distribution of the contribution of different image regions when a certain class of object is recognized. Following the same scheme, we visualize the CAM of different stages and different examples in figure \ref{fig:stage_cam} and figure \ref{fig:cam}. As mentioned before, GAP with the spatial attention layers (SA) distributes the focus of the model, thus covers more details of the image. Since the feature map from each stage of the proposed model has a deeply supervised branch, we can visualize the CAM for different stages given an input image. Figure \ref{fig:stage_cam} compares the CAM using GAP with or without SA for each stage. GAP with SA shows a pattern of less concentrated attention in general no matter how deep the feature map is. The activations from stage 1 and 2 represents some low-level and intermediate-level features. However, in the stage 3 and 4, the receptive field of the activations can cover the entire width of the image. Therefore, the highlighted area on CAM is more elongated. But still, GAP with SA is more scattered along the vertical axis.

Since the CAM of stage 2 shows a more clear comparison, we study some failure cases of GAP under cross-camera matching and visualize the corresponding CAM for stage 2 in figure \ref{fig:cam}.
%In \cite{CAM}, Zhou et al. used a heat map to illustrate the distribution of the contribution of different image regions when a certain class of object is recognized. Following the same scheme, we visualize some class activation maps for four examples in figure \ref{fig:cam}.
The first test image is from the same camera view as the query image and the second test image is from another camera view. In each example, GAP and GAP with SA are both able to recognize the image from the same camera view. However, GAP fails to identify the person from another camera view while GAP with SA still recognizes it. The second and third rows show the CAM of the model with or without the spatial attention layer respectively. Because of lack of spatial relation, the attention of the model without SA is restricted to some certain region of the image. Therefore, the model tends to focus on the most discriminative part of the person, for example, the handbag of the lady in the first example. However, this can be harmful if the expected feature is missing in another camera view. From the class activation map at the lower right corner of the first example, we can see that even though the model tries to find the handbag on the image, the patterns it found look not so convincing to make the claim that it is the same person since the red color of the handbag from the front view is absent. Therefore it fails to recognize her.
%The original images of the same person from different camera views are at the first row. The second row shows the class activation map of the model without the spatial attention layer. The model without the spatial attention can recognize the first image (the left one) but fail on the other image in each example. Because of lack of spatial relation, the attention of the model is restricted to some certain region of the image. Therefore, the model tends to focus on the most discriminative part of the person, for example, the face of the lady in the first example. However, this can be harmful if the expected feature is missing under another camera view. From the class activation map of the second image of the lady in red, we can see that even though the model tries to find some face patterns on the image, the pattern it found looks not so convincing to make the claim that it is the same person. Therefore it fails to recognize her.
This is, however, not the case for GAP with SA since it covers a lot more different regions on the image. For example, the edge of the skirt and the haircut are also involved during the recognition. These features may not be discriminative enough alone, but the combination of them can alter the decision. This redundancy equips the model with robustness. The image (b) only shows the back of the lady, but the model with spatial attention can still recognize her correctly based on the edge of the skirt and the haircut. The same thing happened also in other examples.

\begin{table}[!ht]
\begin{center}
\begin{tabular}{c|cc|cc}
\hline \hline
Metric(\%) & SA1 & SA2 & R-1 & mAP \\ \hline
 & no & - & 84.8 & 67.0 \\
\multirow{-2}{*}{Backbone} & yes & - & {\color[HTML]{3166FF} \textbf{87.4}} & {\color[HTML]{3166FF} \textbf{68.1}} \\ \hline
 & no & no & 86.2 & 68.3 \\
 & no & yes & 86.6 & 68.8 \\
\multirow{-3}{*}{Backbone + DS} & yes & yes & {\color[HTML]{3166FF} \textbf{87.1}} & {\color[HTML]{3166FF} \textbf{69.4}} \\ \hline
 & no & no & 91.7 & 77.0 \\
 & no & yes & {\color[HTML]{333333} 92.4} & {\color[HTML]{333333} 78.2} \\
\multirow{-3}{*}{Backbone + P + DS} & yes & yes & {\color[HTML]{3166FF} \textbf{92.6}} & {\color[HTML]{3166FF} \textbf{78.3}} \\ \hline \hline
\end{tabular}
\end{center}
\caption{The results are evaluated on Market-1501. P represents part classifiers; DS represents deep supervision; SA1 represents the spatial attention layer on the backbone; SA2 represents the spatial attention layer on the deeply supervised branches.}
\label{table:whrclassifier}
\end{table}

\subsection{Ablation Study} \label{ablation}
We show in this section that the spatial attention improves the performance consistently. The ablation study is conducted on the Market-1501 dataset because it gives relatively stable results due to its large scale. Random erasing and re-ranking are not used here. All  models are trained until convergence. For the baseline, we use the blue part in figure \ref{fig:train_arch}. It is basically just a normal ResNet-50 with one additional convolution layer at the very end. This convolution layer is to shrink the 2048-channel feature map to a 256-channel feature map. This baseline model gives 84.8\% R-1 and 67.0\% mAP. Based on this, we evaluate the performance gain of the spatial attention layers. The results are shown in the table \ref{table:whrclassifier}. DS denotes the deep supervision branch in figure \ref{fig:train_arch}. P denotes the usage of part classifiers. SA1 and SA2 denote the spatial attention layer for the backbone and the deeply supervised branches respectively. The models with the spatial attention layers show a consistent improvement to the ones without it. This demonstrates the effectiveness of our proposed spatial attention.

%Based on this, we evaluate the performance gain of partitioning \cite{pcb}, deep supervision \cite{dare} and the spatial attention. The comparisons are summarized in table \ref{table:whrclassifier}. Partitioning means we use exactly the same backbone as in figure \ref{fig:train_arch}, but the spatial attention layer is discharged and only $L_F$ is computed as the total loss. This structure gives 7.5\% and 11\% improvement to R-1 and mAP respectively. The second experiment is for deep supervision, where only the green branches (without spatial attention) are added to the baseline model. This improves the R-1 and mAP by 1.4\% and 1.3\%. The last experiment is for the spatial attention. We insert one spatial attention layer before the GAP in the baseline model. This gives 1.2\% and 0.2\% improvement to R-1 and mAP.

Another experiment is shown in table \ref{table:sa} and compares the proposed architecture with or without the spatial attention layer across four datasets. Random erasing and horizontal flipping are used as data augmentation. The other settings are the same as the previous experiment. From table \ref{table:sa}, a consistent improvement for models with spatial attention can be observed again. On the CUHK03 labeled dataset, the model equipped with spatial attention outperforms the attention-free model by a large margin of 2.4\% for the rank-1 accuracy. The largest gain on mAP of 1.8\% is for DukeMTMC-ReID.

%the results of the gets rid of all the spatial attention layers from the original model with other settings being the same. Table \ref{table:sa} summaries the performances on all four datasets.  Re-ranking is not applied here. All models are trained until convergence. A consistent improvement for models with spatial attention over ones without it. On CUHK03 labeled dataset, the model equipped with spatial attention outperforms the attention-free model with a large margin by 2.4\% for the rank-1 accuracy. The largest gain on mAP of 1.8\% is from DukeMTMC-ReID.

\begin{table}[!ht]
\begin{center}
\begin{tabular}{c|c|c|c}
\hline \hline
\multicolumn{2}{c|}{{\color[HTML]{333333} Metrics(\%)}} & +SA & -SA \\ \hline
 & R-1 & {\color[HTML]{3166FF} \textbf{93.3}} & 92.5 \\
\multirow{-2}{*}{Market-1501} & mAP & {\color[HTML]{3166FF} \textbf{81.7}} & 79.6 \\ \hline
 & R-1 & {\color[HTML]{3166FF} \textbf{84.3}} & 83.4 \\
\multirow{-2}{*}{DukeMTMC-ReID} & mAP & {\color[HTML]{3166FF} \textbf{72.1}} & 70.3 \\ \hline
 & R-1 & {\color[HTML]{3166FF} \textbf{64.3}} & 61.9 \\
\multirow{-2}{*}{CUHK03 labeled} & mAP & {\color[HTML]{3166FF} \textbf{61.4}} & 60.3 \\ \hline
 & R-1 & {\color[HTML]{3166FF} \textbf{58.9}} & 57.8 \\
\multirow{-2}{*}{CUHK03 detected} & mAP & {\color[HTML]{3166FF} \textbf{57.5}} & 55.8 \\ \hline \hline
\end{tabular}
\end{center}
\caption{The proposed architecture from section \ref{net} is used here. The models with the spatial attention shows a consistent improvement over the ones without it. All the models are trained with random erasing until convergence. Re-ranking is not applied.}
\label{table:sa}
\end{table}

\section{Conclusion}
We propose a parameter-free spatial attention layer to incorporate the spatial relation among the activations on the same feature map for GAP. It is computationally efficient and produces consistent improvement over the model without it. The mechanism behind this module is studied both analytically and empirically in order to understand the behavior. We expect a broader range of applications for it.

\bibliographystyle{plain} % We choose the "plain" reference style
\bibliography{noafe_bib} % Entries are in the "refs.bib" file

\begin{thebibliography}{10}

\bibitem{attention_origin}
Dzmitry Bahdanau, Kyunghyun Cho, and Yoshua Bengio.
\newblock Neural machine translation by jointly learning to align and
  translate.
\newblock {\em arXiv preprint arXiv:1409.0473}, 2014.

\bibitem{minmaxconv}
Michael Blot, Matthieu Cord, and Nicolas Thome.
\newblock Max-min convolutional neural networks for image classification.
\newblock In {\em Image Processing (ICIP), 2016 IEEE International Conference
  on}, pages 3678--3682. IEEE, 2016.

\bibitem{dpfl}
Yanbei Chen, Xiatian Zhu, and Shaogang Gong.
\newblock Person re-identification by deep learning multi-scale
  representations.
\newblock In {\em IEEE International Conference on Computer Vision Workshop},
  pages 2590--2600, 2017.

\bibitem{fmn}
Guodong Ding, Salman Khan, Zhenmin Tang, and Fatih Porikli.
\newblock Let features decide for themselves: Feature mask network for person
  re-identification.
\newblock 2017.

\bibitem{wildcat}
Thibaut Durand, Taylor Mordan, Nicolas Thome, and Matthieu Cord.
\newblock Wildcat: Weakly supervised learning of deep convnets for image
  classification, pointwise localization and segmentation.
\newblock In {\em IEEE Conference on Computer Vision and Pattern Recognition
  (CVPR 2017)}, volume~2, 2017.

\bibitem{mantra}
Thibaut Durand, Nicolas Thome, and Matthieu Cord.
\newblock Mantra: minimum maximum latent structural svm for image
  classification and ranking.
\newblock In {\em Proceedings of the IEEE International Conference on Computer
  Vision}, pages 2713--2721, 2015.

\bibitem{weldon}
Thibaut Durand, Nicolas Thome, and Matthieu Cord.
\newblock Weldon: Weakly supervised learning of deep convolutional neural
  networks.
\newblock In {\em Proceedings of the IEEE Conference on Computer Vision and
  Pattern Recognition}, pages 4743--4752, 2016.

\bibitem{hpm}
Yang Fu, Yunchao Wei, Yuqian Zhou, Honghui Shi, Gao Huang, Xinchao Wang,
  Zhiqiang Yao, and Thomas Huang.
\newblock Horizontal pyramid matching for person re-identification.
\newblock 2018.

\bibitem{senet}
Jie Hu, Li~Shen, and Gang Sun.
\newblock Squeeze-and-excitation networks.
\newblock {\em arXiv preprint arXiv:1709.01507}, 7, 2017.

\bibitem{deepsupervision}
Chen-Yu Lee, Saining Xie, Patrick Gallagher, Zhengyou Zhang, and Zhuowen Tu.
\newblock Deeply-supervised nets.
\newblock In {\em Artificial Intelligence and Statistics}, pages 562--570,
  2015.

\bibitem{cuhk03}
Wei Li, Rui Zhao, Tong Xiao, and Xiaogang Wang.
\newblock Deepreid: Deep filter pairing neural network for person
  re-identification.
\newblock In {\em CVPR}, 2014.

\bibitem{atten_reid_harm}
Wei Li, Xiatian Zhu, and Shaogang Gong.
\newblock Harmonious attention network for person re-identification.
\newblock In {\em CVPR}, volume~1, page~2, 2018.

\bibitem{MIL}
Weixin Li and Nuno Vasconcelos.
\newblock Multiple instance learning for soft bags via top instances.
\newblock In {\em Proceedings of the ieee conference on computer vision and
  pattern recognition}, pages 4277--4285, 2015.

\bibitem{GAP}
Min Lin, Qiang Chen, and Shuicheng Yan.
\newblock Network in network.
\newblock {\em arXiv preprint arXiv:1312.4400}, 2013.

\bibitem{isthere}
Maxime Oquab, L{\'e}on Bottou, Ivan Laptev, and Josef Sivic.
\newblock Is object localization for free?-weakly-supervised learning with
  convolutional neural networks.
\newblock In {\em Proceedings of the IEEE Conference on Computer Vision and
  Pattern Recognition}, pages 685--694, 2015.

\bibitem{parizi2014automatic}
Sobhan~Naderi Parizi, Andrea Vedaldi, Andrew Zisserman, and Pedro Felzenszwalb.
\newblock Automatic discovery and optimization of parts for image
  classification.
\newblock {\em arXiv preprint arXiv:1412.6598}, 2014.

\bibitem{duke2}
Ergys Ristani, Francesco Solera, Roger Zou, Rita Cucchiara, and Carlo Tomasi.
\newblock Performance measures and a data set for multi-target, multi-camera
  tracking.
\newblock In {\em European Conference on Computer Vision workshop on
  Benchmarking Multi-Target Tracking}, 2016.

\bibitem{crelu}
Wenling Shang, Kihyuk Sohn, Diogo Almeida, and Honglak Lee.
\newblock Understanding and improving convolutional neural networks via
  concatenated rectified linear units.
\newblock In {\em International Conference on Machine Learning}, pages
  2217--2225, 2016.

\bibitem{atten_reid_harm3}
Chen Shen, Guo-Jun Qi, Rongxin Jiang, Zhongming Jin, Hongwei Yong, Yaowu Chen,
  and Xian-Sheng Hua.
\newblock Sharp attention network via adaptive sampling for person
  re-identification.
\newblock {\em arXiv preprint arXiv:1805.02336}, 2018.

\bibitem{pcb1}
Yang Shen, Weiyao Lin, Junchi Yan, Mingliang Xu, Jianxin Wu, and Jingdong Wang.
\newblock Person re-identification with correspondence structure learning.
\newblock In {\em Proceedings of the IEEE International Conference on Computer
  Vision}, pages 3200--3208, 2015.

\bibitem{pcb3}
Hailin Shi, Xiangyu Zhu, Shengcai Liao, Zhen Lei, Yang Yang, and Stan~Z Li.
\newblock Constrained deep metric learning for person re-identification.
\newblock {\em arXiv preprint arXiv:1511.07545}, 2015.

\bibitem{atten_reid_harm1}
Jianlou Si, Honggang Zhang, Chun-Guang Li, Jason Kuen, Xiangfei Kong, Alex~C
  Kot, and Gang Wang.
\newblock Dual attention matching network for context-aware feature sequence
  based person re-identification.
\newblock {\em arXiv preprint arXiv:1803.09937}, 2018.

\bibitem{atten_reid_harm5}
Chunfeng Song, Yan Huang, Wanli Ouyang, and Liang Wang.
\newblock Mask-guided contrastive attention model for person re-identification.
\newblock In {\em Proceedings of the IEEE Conference on Computer Vision and
  Pattern Recognition}, pages 1179--1188, 2018.

\bibitem{dropout}
Nitish Srivastava, Geoffrey Hinton, Alex Krizhevsky, Ilya Sutskever, and Ruslan
  Salakhutdinov.
\newblock Dropout: A simple way to prevent neural networks from overfitting.
\newblock {\em Journal of Machine Learning Research}, 15:1929--1958, 2014.

\bibitem{pcb}
Yifan Sun, Liang Zheng, Yi~Yang, Qi~Tian, and Shengjin Wang.
\newblock Beyond part models: Person retrieval with refined part pooling.
\newblock {\em arXiv preprint arXiv:1711.09349}, 2017.

\bibitem{BOW}
Manik Varma and Andrew Zisserman.
\newblock A statistical approach to texture classification from single images.
\newblock {\em International journal of computer vision}, 62(1-2):61--81, 2005.

\bibitem{mancs}
Cheng Wang, Qian Zhang, Chang Huang, Wenyu Liu, and Xinggang Wang.
\newblock Mancs: A multi-task attentional network with curriculum sampling for
  person re-identification.
\newblock In {\em Proceedings of the European Conference on Computer Vision
  (ECCV)}, pages 365--381, 2018.

\bibitem{dare}
Yan Wang, Lequn Wang, Yurong You, Xu~Zou, Vincent Chen, Serena Li, Gao Huang,
  Bharath Hariharan, and Kilian~Q Weinberger.
\newblock Resource aware person re-identification across multiple resolutions.
\newblock In {\em Proceedings of the IEEE Conference on Computer Vision and
  Pattern Recognition}, pages 8042--8051, 2018.

\bibitem{deepsuper_reid}
Yan Wang, Lequn Wang, Yurong You, Xu~Zou, Vincent Chen, Serena Li, Gao Huang,
  Bharath Hariharan, and Kilian~Q Weinberger.
\newblock Resource aware person re-identification across multiple resolutions.
\newblock In {\em Proceedings of the IEEE Conference on Computer Vision and
  Pattern Recognition}, pages 8042--8051, 2018.

\bibitem{oimloss}
Tong Xiao, Shuang Li, Bochao Wang, Liang Lin, and Xiaogang Wang.
\newblock Joint detection and identification feature learning for person
  search.
\newblock 2017.

\bibitem{atten_reid_harm2}
Jing Xu, Rui Zhao, Feng Zhu, Huaming Wang, and Wanli Ouyang.
\newblock Attention-aware compositional network for person re-identification.
\newblock {\em arXiv preprint arXiv:1805.03344}, 2018.

\bibitem{aacn}
Jing Xu, Rui Zhao, Feng Zhu, Huaming Wang, and Wanli Ouyang.
\newblock Attention-aware compositional network for person re-identification.
\newblock {\em arXiv preprint arXiv:1805.03344}, 2018.

\bibitem{atten_reid_harm4}
Shuangjie Xu, Yu~Cheng, Kang Gu, Yang Yang, Shiyu Chang, and Pan Zhou.
\newblock Jointly attentive spatial-temporal pooling networks for video-based
  person re-identification.
\newblock {\em arXiv preprint arXiv:1708.02286}, 2017.

\bibitem{market1501}
Liang Zheng, Liyue Shen, Lu~Tian, Shengjin Wang, Jingdong Wang, and Qi~Tian.
\newblock Scalable person re-identification: A benchmark.
\newblock In {\em Computer Vision, IEEE International Conference on}, 2015.

\bibitem{pcb2}
Wei-Shi Zheng, Xiang Li, Tao Xiang, Shengcai Liao, Jianhuang Lai, and Shaogang
  Gong.
\newblock Partial person re-identification.
\newblock In {\em Proceedings of the IEEE International Conference on Computer
  Vision}, pages 4678--4686, 2015.

\bibitem{duke1}
Zhedong Zheng, Liang Zheng, and Yi~Yang.
\newblock Unlabeled samples generated by gan improve the person
  re-identification baseline in vitro.
\newblock In {\em Proceedings of the IEEE International Conference on Computer
  Vision}, 2017.

\bibitem{pan}
Zhedong Zheng, Liang Zheng, and Yi~Yang.
\newblock Pedestrian alignment network for large-scale person
  re-identification.
\newblock {\em IEEE Transactions on Circuits and Systems for Video Technology},
  2018.

\bibitem{reranking}
Zhun Zhong, Liang Zheng, Donglin Cao, and Shaozi Li.
\newblock Re-ranking person re-identification with k-reciprocal encoding.
\newblock In {\em Computer Vision and Pattern Recognition (CVPR), 2017 IEEE
  Conference on}, pages 3652--3661. IEEE, 2017.

\bibitem{random_erase}
Zhun Zhong, Liang Zheng, Guoliang Kang, Shaozi Li, and Yi~Yang.
\newblock Random erasing data augmentation.
\newblock {\em arXiv preprint arXiv:1708.04896}, 2017.

\bibitem{CAM}
Bolei Zhou, Aditya Khosla, Agata Lapedriza, Aude Oliva, and Antonio Torralba.
\newblock Learning deep features for discriminative localization.
\newblock In {\em Proceedings of the IEEE Conference on Computer Vision and
  Pattern Recognition}, pages 2921--2929, 2016.

\end{thebibliography}

\end{document}